%% file: root.tex
\documentclass[10pt,conference,a4paper]{IEEEtran}
\usepackage{cite}
\usepackage{amsmath,amssymb,amsfonts}
\usepackage{mathrsfs}
\usepackage{algorithmic}
\usepackage{graphicx}
\usepackage{textcomp}
\usepackage{xcolor}
\usepackage{mathtools}
\usepackage{booktabs}
\usepackage[colorlinks=false]{hyperref}
\def\BibTeX{{\rm B\kern-.05em{\sc i\kern-.025em b}\kern-.08em
    T\kern-.1667em\lower.7ex\hbox{E}\kern-.125emX}}
    
\usepackage{glossaries}
\usepackage{balance}
\usepackage[caption=false]{subfig}

\DeclarePairedDelimiter{\norma}{\lVert}{\rVert}
\definecolor{mat_orange}{HTML}{FF7F0E}
\definecolor{mat_blue}{HTML}{1F77B4}

\newacronym{dfdc}{DFDC}{DeepFake Detection Challenge}
\newacronym{ff++}{FF++}{FaceForensics++}
\newacronym{cnn}{CNN}{Convolutional Neural Network}
\newacronym{lstm}{LSTM}{Long Short-Term Memory}
\newacronym{auc}{AUC}{area under the curve}

\begin{document}

\title{Video Face Manipulation Detection Through Ensemble of CNNs\\
}

\makeatletter
\newcommand{\linebreakand}{%
  \end{@IEEEauthorhalign}
  \hfill\mbox{}\par
  \mbox{}\hfill\begin{@IEEEauthorhalign}
}
\makeatother

\author{\IEEEauthorblockN{Nicol\`o~Bonettini}
\IEEEauthorblockA{\textit{DEIB} \\
\textit{Politecnico di Milano}\\
Milano, Italy \\
nicolo.bonettini@polimi.it}
\and
\IEEEauthorblockN{Edoardo~Daniele~Cannas}
\IEEEauthorblockA{\textit{DEIB} \\
\textit{Politecnico di Milano}\\
Milano, Italy \\
edoardodaniele.cannas@polimi.it}
\and
\IEEEauthorblockN{Sara~Mandelli}
\IEEEauthorblockA{\textit{DEIB} \\
\textit{Politecnico di Milano}\\
Milano, Italy \\
sara.mandelli@polimi.it}
\linebreakand
\IEEEauthorblockN{Luca~Bondi}
\IEEEauthorblockA{\textit{DEIB} \\
\textit{Politecnico di Milano}\\
Milano, Italy \\
luca.bondi@polimi.it}
\and
\IEEEauthorblockN{Paolo~Bestagini}
\IEEEauthorblockA{\textit{DEIB} \\
\textit{Politecnico di Milano}\\
Milano, Italy \\
paolo.bestagini@polimi.it}
\and
\IEEEauthorblockN{Stefano~Tubaro}
\IEEEauthorblockA{\textit{DEIB} \\
\textit{Politecnico di Milano}\\
Milano, Italy \\
stefano.tubaro@polimi.it}
}

\maketitle

\begin{abstract}
In the last few years, several techniques for facial manipulation in videos have been successfully developed and made available to the masses (i.e., FaceSwap, deepfake, etc.). 
These methods enable anyone to easily edit faces in video sequences with incredibly realistic results and a very little effort.
Despite the usefulness of these tools in many fields, if used maliciously, they can have a significantly bad impact on society (e.g., fake news spreading, cyber bullying through fake revenge porn).
The ability of objectively detecting whether a face has been manipulated in a video sequence is then a task of utmost importance.

In this paper, we tackle the problem of face manipulation detection in video sequences targeting modern facial manipulation techniques.
In particular, we study the ensembling of different trained \gls{cnn} models.
In the proposed solution, different models are obtained starting from a base network (i.e., EfficientNetB4) making use of two different concepts: (i) attention layers; (ii) siamese training.
We show that combining these networks leads to promising face manipulation detection results on two publicly available datasets with more than 119000 videos.
\end{abstract}


\input{introduction}
\input{sota}
\input{method}

\input{experiments}
\input{results}
\input{conclusions}

\bibliographystyle{IEEEtran}
\balance
\bibliography{biblio}

\end{document}

%% file: introduction.tex
\section{Introduction}
\label{sec:introduction}
Over the past few years, huge steps forward in the field of automatic video editing techniques have been made.
In particular, great interest has been shown towards methods for facial manipulation \cite{Zollhofer2018}.
Just to name an example, it is nowadays possible to easily perform facial reenactment, i.e., transferring the facial expressions from one video to another one \cite{Thies2016face2face, Thies2019deferred}.
This enables to change the identity of a speaker with very little effort.

Systems and tools for facial manipulations are now so advanced that even users without any previous experience in photo retouching and digital arts can use them.
Indeed, code and libraries that work in an almost automatic fashion are more and more often made available to the public for free \cite{Deepfake, Faceswap}.
On one hand, this technological advancement opens the door to new artistic possibilities (e.g., movie making, visual effect, visual arts, etc.).
On the other hand, unfortunately, it also eases the generation of video forgeries by malicious users.

Fake news spreading and revenge porn are just a few of the possible malicious applications of advanced facial manipulation technology in the wrong hands.
As the distribution of these kinds of manipulated videos indubitably leads to serious and dangerous consequences (e.g., diminished trust in media, targeted opinion formation, cyber bullying, etc.), the ability of detecting whether a face has been manipulated in a video sequence is becoming of paramount importance \cite{Agarwal_2019_CVPR_Workshops}.

Detecting whether a video has been modified is not a novel issue per se.
Multimedia forensics researchers have been working on this topic since many years, proposing different kinds of solutions to different problems \cite{Rocha2011a, Milani2012a, Stamm2013}.
For instance, in \cite{bestagini2016codec, Padin2020} the authors focus on studying the coding history of videos.
The authors of \cite{Bestagini2013, Damiano2019} focus on localizing copy-move forgeries with block-based or dense techniques.
In \cite{Stamm2012, Gironi2014}, different methods are proposed to detect frame duplication or deletion.

\begin{figure*}[t]
	\centering	\includegraphics[width=\textwidth]{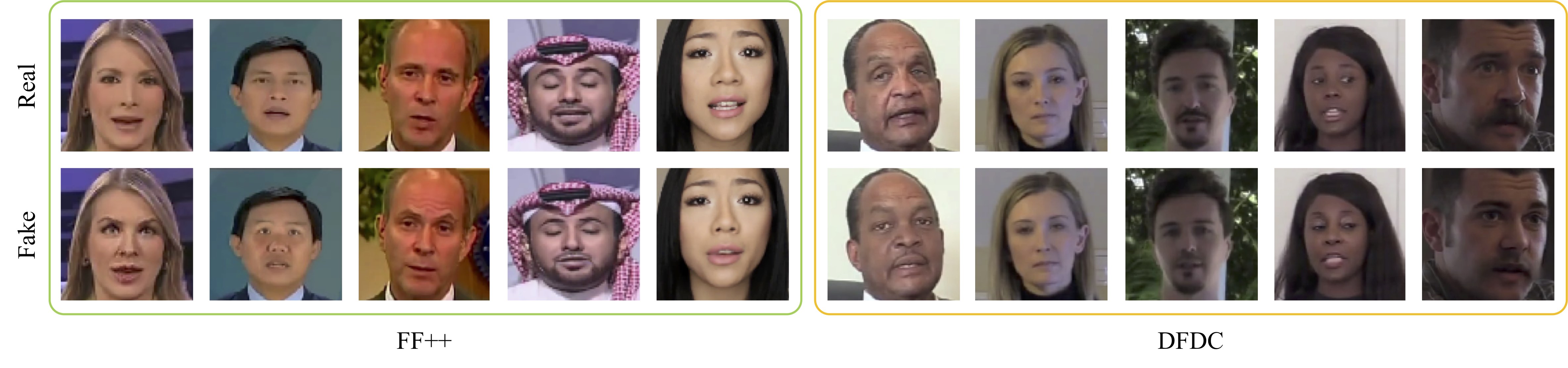}
	\caption{Sample faces extracted from \gls{ff++} and \gls{dfdc} datasets. For each pristine face, we show a corresponding fake sample generated from it.} 
	\label{fig:sample_faces}
\end{figure*}

All the above-mentioned methods work according to a common principle: each non-reversible operation leaves a peculiar footprint that can be exposed to detect the specific editing.
However, forensics footprints are often very subtle and hard to detect.
This is the case of videos undergoing excessive compression, multiple editing operations at once, or strong downsampling \cite{Milani2012a}.
This is also the case of very realistic forgeries operated through methods that are hard to formally model.
For this reason, modern facial manipulation techniques are very challenging to detect from the forensic perspective \cite{verdoliva2020media}.
As a matter of fact, many different face manipulation techniques exist (i.e., there is not a unique model explaining these forgeries).
Moreover, they often operate on small video regions only (i.e., the face or part of it, and not the full frame).
Finally, these kinds of manipulated videos are typically shared through social platforms that apply resizing as well as coding steps, further hindering classic forensic detectors performance.

In this paper, we tackle the problem of detecting facial manipulation operated through modern solutions.
In particular, we focus on all the manipulation techniques reported in \cite{roessler2019faceforensics} (i.e., deepfakes, Face2Face, FaceSwap and NeuralTextures) and in the Facebook \gls{dfdc} started on Kaggle in December 2019 \cite{dfdc2019}.
Within this context, we study the possibility of using an ensemble of different \gls{cnn} trained models.
We consider EfficientNetB4 \cite{Tan2019efficientnet} and propose a modified version of it obtained by adding an attention mechanism \cite{vaswani2017}.
Moreover, for each network, we investigate two different training strategies, one of which is based on the siamese paradigm.

As one of the big challenges is to be able to run a forensic detector in real-world scenarios, we develop our solution keeping computational complexity at bay.
Specifically, we consider the strong hardware and time constraints imposed by the \gls{dfdc} \cite{dfdc2019}.
This means that the proposed solution must be able to analyze $4\,000$ videos in less than 9 hours using at most a single NVIDIA P100 GPU.
Moreover, the trained models must occupy less than 1GB of disk space.

Evaluation is performed on two disjoint datasets: \gls{ff++} \cite{roessler2019faceforensics}, which has been recently proposed as a public benchmark; \gls{dfdc} \cite{dfdc2019}, which has been released as part of the \gls{dfdc} Kaggle competition.
Fig. \ref{fig:sample_faces} depicts a few examples of faces extracted from the two datasets, reporting pristine and manipulated samples.
Results show that the proposed attention-based modification as well as the siamese training strategy help the ensemble system in outperforming the baseline reported in \gls{ff++} on both datasets.
Moreover, the proposed attention-based solution provides interesting insights on which part of each frame drives face manipulation detection, thus enabling a small step forward towards the explainability of the network results.

The rest of the paper is structured as follows.
Section~\ref{sec:sota} reports a literature review of the latest related work.
Section~\ref{sec:method} reports all the details about the proposed method.
Section~\ref{sec:experiments} details the experimental setup.
Section~\ref{sec:results} collects all the achieved results.
Finally, Section~\ref{sec:conclusions} concludes the paper.

%% file: sota.tex
\section{Related work}\label{sec:sota}
Multiple video forensics techniques have been proposed for a variety of tasks in the last few years \cite{Rocha2011a, Milani2012a, Stamm2013}.
However, since the forensics community has become aware of the potential social risks introduced by the latest facial manipulation techniques, many detection algorithms have been proposed to detect this kind of forgeries \cite{verdoliva2020media}.

Some of the proposed techniques focus on a \gls{cnn}-based frame-by-frame analysis.
For instance, MesoNet is proposed in \cite{mesonet}.
This is a relatively shallow \gls{cnn} with the goal of detecting fake faces.
The authors of \cite{roessler2019faceforensics} have shown that this network is outperformed by XceptionNet retrained on purpose.

Alternative techniques exploit also the temporal evolution of video frames through \gls{lstm} analysis.
This is the case of \cite{DBLP:journals/corr/abs-1812-08685} and \cite{Guera2019}, which first extract a series of frame-based features, and then put them together with a recurrent mechanism.

Other methods leverage specific processing traces. 
This is the case of \cite{li2019exposing}, where the authors exploit the fact that deepfake donor faces are warped in order to realistically stick to the host video.
They therefore propose a detector that captures warping traces.

In order to overcome the limitation of pixel analysis, other techniques are based on a semantic analysis of the frames.
In \cite{Yang2019}, a technique that learns to distinguish natural and fake head pose is proposed.
Conversely, the authors of \cite{matern2019exploiting} focus on inconsistent lighting effects.
Alternatively, \cite{Li2018} reports a methodology based on eye blinking analysis.
Indeed, the first generation of deepfake videos was showing some eye artifacts that could be captured with this method.
Unfortunately, the more the manipulation techniques produce realistic results, the less semantic methods work. 

Finally, other techniques provide additional localization information.
The authors of \cite{DBLP:journals/corr/abs-1906-06876} propose a multi-task learning method that provides a detection score together with a segmentation mask.
Alternatively, in \cite{dang2019detection}, an attention mechanism is proposed.

Inspired by the state of the art, in this paper we focus on network ensembles, proposing a solution that works on multiple datasets and is sufficiently lightweight according to \gls{dfdc} competition rules \cite{dfdc2019}.

%% file: method.tex
\section{Proposed method}
\label{sec:method}
In this section, we describe our proposed method for video face manipulation detection, i.e., given a video frame, to detect whether faces are real (pristine) or fake.

The proposed method is based on the concept of ensembling.
Indeed, it is well-known that model ensembling may lead to better prediction performance.
We therefore focus on investigating whether and how it is possible to train different \gls{cnn}-based classifiers to capture different high-level semantic information that complement one another, thus positively contributing to the ensemble for this specific problem.

To do so, we consider as starting point the EfficientNet family of models, proposed in \cite{Tan2019efficientnet} as a novel approach for the automatic scaling of \gls{cnn}s.
This set of architectures achieves better accuracy and efficiency with respect to other state-of-the-art \gls{cnn}s, and actually revealed to be very useful to fulfil hardware and time constraints imposed by \gls{dfdc}.
Given an EfficientNet architecture, we propose to follow two paths to make the model beneficial for the ensambling.
On one hand, we propose to include an attention mechanism, which also provides the analyst with a method to infer which portion of the investigated video is more informative for the classification process.
On the other hand, we investigate how siamese training strategies can be included into the learning process for extrapolating additional information about the data.

In the following, more details are provided about EfficientNet architecture with the proposed attention mechanism and the network training strategies.

\subsection{EfficientNet and attention mechanism}
Among the family of EfficientNet models, we choose the EfficientNetB4 as the baseline for our work, motivated by the good trade-off offered by this architecture in terms of dimensions (i.e., number of parameters), run time (i.e., FLOPS cost) and classification performance.
As reported in \cite{Tan2019efficientnet}, with 19 millions of parameters and 4.2 billions of FLOPS, EfficientNetB4 reaches the 83.8\% top-1 accuracy on the ImageNet \cite{imagenet_cvpr09} dataset. 
On the same dataset, XceptionNet, used as face manipulation detection baseline method by the authors of \cite{roessler2019faceforensics}, reaches the 79\% top-1 accuracy at the expense of 23 millions parameters and 8.4 billions FLOPS.

\begin{figure}[t]
	\centering	\includegraphics[width=\columnwidth]{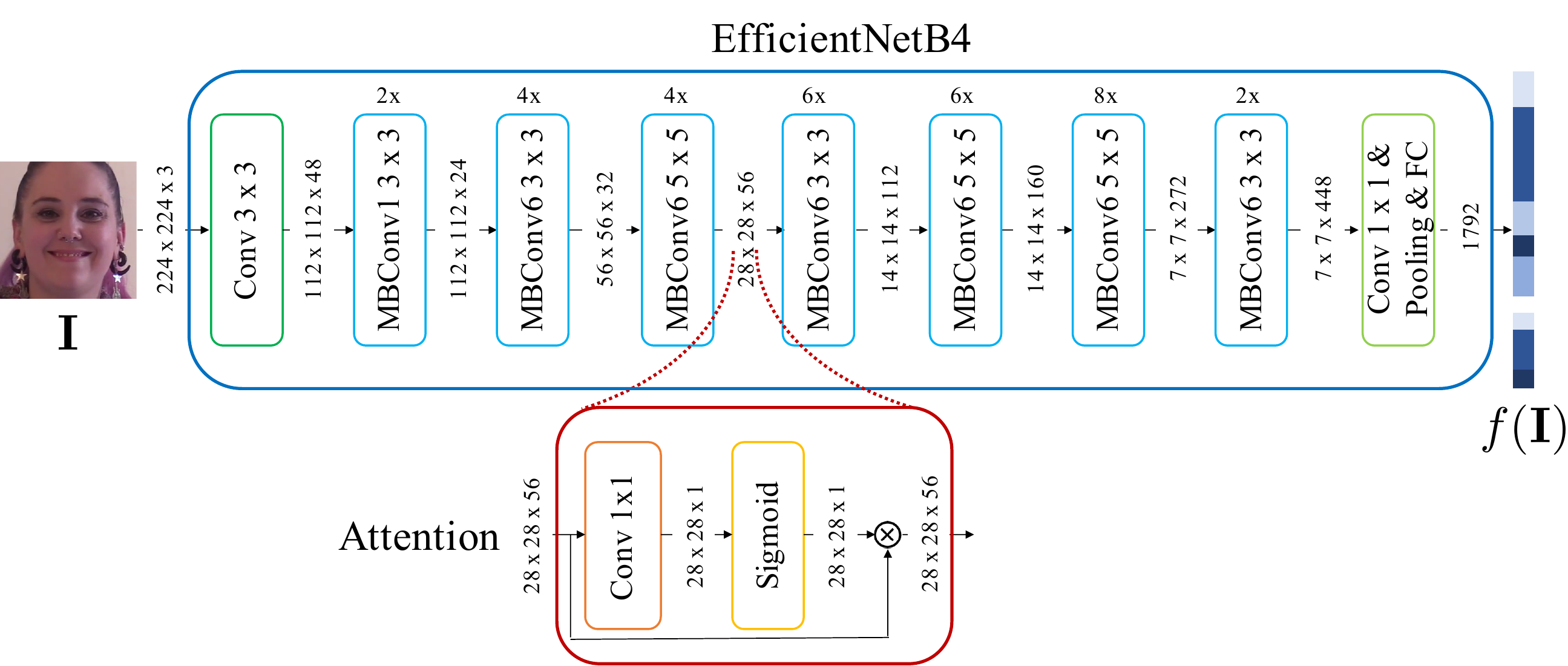}
	\caption{Blue block: EfficientNetB4 model. If the red block is embedded into the network, an attention mechanism is included in the model, defining the proposed EfficientNetB4Att architecture.}
	\label{fig:efficientnetB4att}
\end{figure}

EfficientNetB4 architecture is represented within the blue block in Fig.~\ref{fig:efficientnetB4att}, where all layers are defined using the same nomenclature introduced in \cite{Tan2019efficientnet}. 

The input to the network is a squared color image $\mathbf{I}$, i.e., in our experiments, the face extracted from a video frame.
As a matter of fact, authors of \cite{roessler2019faceforensics} recommend to track face information instead of using the full frame as input to the network for increasing the classification accuracy.
Moreover, faces can be easily extracted from frames using any of the widely available face detectors proposed in the literature \cite{Zhang2016,blazeface2019}.
The network output is a feature vector of $1792$ elements, defined as $f(\mathbf{I})$.
The final score related to the face is the result of a classification layer.

The proposed variant of the standard EfficientNetB4 architecture
is inspired by the several contributions in the natural language processing and computer vision fields that make use of attention mechanisms.
Works such as the transformer \cite{vaswani2017} and residual attention networks \cite{Wang_2017_residual_attention} show how it is possible for a neural network to learn which part of its input (being an image or a sequence of words) is more relevant for accomplishing the task at hand.
In the context of video deepfake detection, it would be of great benefit to discover which portion of the input gave the network more information for its decision making process.
We thus explicitly implement an attention mechanism similar to the one already exploited by the EfficientNet itself, as well as to the self-attention mechanisms presented in \cite{Hu2018, dang2019detection}:
\begin{enumerate}
    \item we select the feature maps extracted by the EfficientNetB4 up to a certain layer, chosen such that these features provide sufficient information on the input frame without being too detailed or, on the contrary, too unrefined. To this purpose, we select the output features at the third MBConv block which have size $28 \times 28 \times 56$;
    \item we process the feature maps with a single convolutional layer with kernel size 1 followed by a Sigmoid activation function to obtain a single attention map;
    \item we multiply the attention map for each of the feature maps at the selected layer. 
\end{enumerate}
For clarity's sake, the attention-based module is depicted in the red block of Fig.~\ref{fig:efficientnetB4att}.

On one hand, this simple mechanism enables the network to focus only on the most relevant portions of the feature maps, on the other hand it provides us with a deeper insight on which parts of the input the network assumes as the most informative. 
Indeed, the obtained attention map can be easily mapped to the input sample, highlighting which elements of it have been given more importance by the network. 
The result of the attention block is finally processed by the remaining layers of EfficientNetB4. 
The whole training procedure can be executed end-to-end, and we call the resulting network EfficientNetB4Att.

\subsection{Network training}
\label{subsec:network_training}
We train each model according to two different training paradigms: (i) end-to-end, and; (ii) siamese.
The former represents a more classical training strategy, also used as evaluation metrics in the contest of \gls{dfdc}.
The latter aims at exploiting the generalization capabilities offered by the networks in order to obtain a feature descriptor that privileges the similarity between samples belonging to the same class.
The ultimate goal is to learn a representation in the encoding space of the network's layers that well separates samples (i.e., faces) of the real and fake class. 

\subsubsection{End-to-end training}
We feed the network with a sample face, and the network returns a face-related score $\hat{y}$. Notice that this score is not passed through a Sigmoid activation function yet.
The weights update is led by the commonly used LogLoss function
\begin{equation}
    L_L=-\frac{1}{N} \sum_{i=1}^{N}\left[y_{i} \log\left( S(\hat{y}_{i})\right)+\left(1-y_{i}\right) \log\left(1- S(\hat{y}_{i})\right)\right],
\end{equation}
where $\hat{y}_i$ represents the $i$-th face score, $y_i \in \{0, 1 \}$ the related face label.
Specifically, label $0$ is associated with faces coming from real pristine videos and label $1$ with fake videos.
$N$ is the total number of faces used for training and $S\left(\cdot\right)$ is the Sigmoid function.

\subsubsection{Siamese training}
Inspired by computer vision works that generate local feature descriptors using \gls{cnn}s, we adopt the triplet margin loss, first proposed in \cite{Wang2014triplet}. 
Recalling that $f(\mathbf{I})$ is the non-linear encoding obtained by the network for an input face $\mathbf{I}$ (see Fig.~\ref{fig:efficientnetB4att}), being $\norma{\cdot}_{2}$ the $L_{2}$ norm, the triplet margin loss is defined as
\begin{equation}
\label{tripletloss}
    L_T = \max(0, \mu + \delta_{+} - \delta_{-}),
\end{equation}
with $\delta_{+}=\norma{f(\mathbf{I}_a)-f(\mathbf{I}_p)}_2$, $\delta_{-}=\norma{f(\mathbf{I}_a)-f(\mathbf{I}_n)}_2$ and $\mu$ is a strictly positive margin. In this case $\mathbf{I}_a,\, \mathbf{I}_p$ and $\mathbf{I}_n$ are, respectively:
\begin{itemize}
    \item $\mathbf{I}_a$ the \textit{anchor} sample (i.e., a real face);
    \item $\mathbf{I}_p$ a \textit{positive} sample, belonging to the same class as $\mathbf{I}_a$ (i.e., another real face);
    \item $\mathbf{I}_n$ a \textit{negative} sample, belonging to a different class than $\mathbf{I}_a$ (i.e., a fake face).
\end{itemize}

We then finalize the training by finetuning a simple classification layer on top of the network, following the end-to-end approach described before.

%% file: experiments.tex
\section{Experiments}\label{sec:experiments}
In this section we report all the details regarding the used datasets and experimental setup.

\subsection{Dataset}
We test the proposed method on two different datasets: \gls{ff++} \cite{roessler2019faceforensics}; \gls{dfdc} \cite{dfdc2019}.

\gls{ff++} is a large-scale facial manipulation dataset generated using automated state-of-the-art video editing methods. In detail, two classical computer graphics approaches are used, i.e., Face2Face \cite{Thies2016face2face} and FaceSwap \cite{Faceswap}, together with two learning-based strategies, i.e., DeepFakes \cite{Deepfake} and NeuralTextures \cite{Thies2019deferred}. 
Every method is applied to $1000$ high quality pristine videos downloaded from YouTube, manually selected to present nearly front-facing subjects without occlusions. All the sequences contain at least $280$ frames.
Eventually, a database of more than $1.8$ million images from $4000$ manipulated videos is built.
In order to simulate a realistic setting, videos are compressed using the H.264 codec. High quality as well as low quality videos are generated using a constant rate quantization parameter equal to $23$ and $40$, respectively.

\gls{dfdc} is the training dataset released for the homologous Kaggle challenge.
It is composed by more than $119\,000$ video sequences, created specifically for this challenge, representing both real and fake videos.
The real videos are sequences of actors taking into account diversity in several axes (gender, skin-tone, age, etc.) recorded with arbitrary backgrounds to bring visual variability.
The fake videos are created starting from the real ones and applying different DeepFake techniques, e.g., different face swap algorithms.
Notice that we do not know the precise algorithms used to generate fake videos, since for the time being the complete dataset (i.e., with the public and private testing sequences and possibly an explanation of the creation procedure) has not been released yet.
The sequence length is roughly $300$ frames, and the classes are strongly unbalanced towards the fake one, counting roughly $100\,000$ fakes and $19\,000$ reals.

\subsection{Networks}
In our experiments, we consider the following networks:
\begin{itemize}
    \item XceptionNet, since it is the best performing model used in \cite{roessler2019faceforensics}, thus being the natural yardstick for our experimental campaign;
    \item EfficentNetB4, as it achieves better accuracy and efficiency than other existing methods \cite{Tan2019efficientnet};    
    \item EfficentNetB4Att, which should discriminate relevant parts of the face sample from irrelevant ones.
\end{itemize}

Each model is trained and tested separately over both the considered datasets. Specifically, regarding \gls{ff++}, we consider only videos generated with constant rate quantization equal to $23$. 
XceptionNet is trained using the same approach of \cite{roessler2019faceforensics}, whereas the two EfficientNet models are trained following the end-to-end as well as the siamese fashion described in Section~\ref{subsec:network_training}.
In doing so, we end up with $4$ trained models: EfficientNetB4 and EfficientNetB4Att which are trained with the classical end-to-end approach, together with EfficientNetB4ST and EfficientNetB4AttST, trained using the siamese strategy. All these EfficientNetB4-derived models can contribute to the final ensembling.

\subsection{Setup}
We adopt a different split policy for each dataset. We split \gls{dfdc} according to its folder structure, using the first $35$ folders for training, folders from $36$ to $40$ for validation and the last $10$ folders for testing.
Regarding \gls{ff++}, we use a similar split as in \cite{roessler2019faceforensics} selecting $720$ videos for training, $140$ for validation and $140$ for test from the pool of original sequences taken from YouTube. The corresponding fake videos are assigned to the same split. All the results are shown on the test sets.

\begin{figure}[t]
	\centering	\includegraphics[width=\columnwidth]{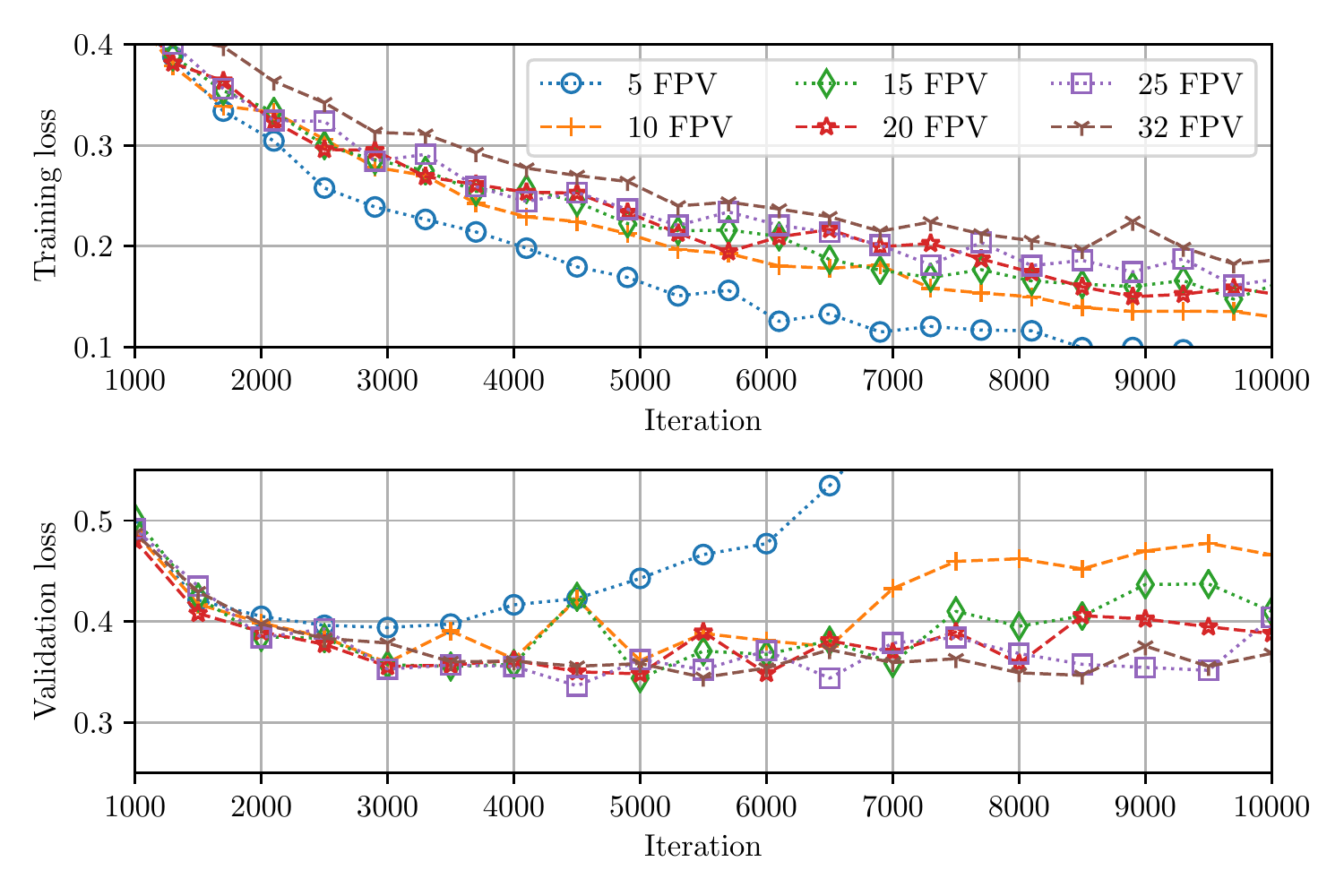}
	\caption{Training and validation loss curves for XceptionNet on \gls{ff++}, while varying the number of frames per video (FPV).}
	\label{fig:figura_overfitting_xception}
\end{figure}

In our experiments, we only consider a limited number of frames for each video.
In training phase, this choice is motivated by two main considerations: (i) when using a really small amount of frames per video, there is a strong tendency to overfit; (ii) increasing the number of frames does not improve performances in a justifiable manner. 
This phenomenon can be noticed in Fig.~\ref{fig:figura_overfitting_xception}, which reports training and validation losses as a function of training iterations, selecting a variable amount of frames per video. It is worth noting that the minimum validation loss does not improve selecting $15$ frames per video instead of $32$, however choosing $32$ frames per video helps to prevent overfitting.
For testing, we should also take into account 
the hardware and time constraints imposed by the \gls{dfdc} challenge. 
With this in mind,
we limit the number of analyzed frames from each sequence to $32$ for both training and testing phases.
Even in this setting, the dimensions of the datasets remain remarkable: for the \gls{ff++}, we end up with roughly $1.6$ million images, while for the \gls{dfdc} with $3.4$ million frames.

In this perspective, we can further reduce the amount of data processed by the networks by recalling that not all the frame information is useful for the deepfake detection process \cite{roessler2019faceforensics}. 
Indeed, we can mainly focus our analysis on the region where the face of the subject is located. Consequently, as a pre-processing step, we extract from each frame the faces of the scene subjects using the BlazeFace extractor \cite{blazeface2019}, that, in our experiments, proved to be faster than the MTCNN detector \cite{Zhang2016} used by the authors of \cite{roessler2019faceforensics}. 
In case more than one face is detected, we keep the face with the best confidence score. The resulting input for the networks is the squared color image $\mathbf{I}$ introduced in section \ref{sec:method}, of size $224\times224$ pixel. 

During training and validation, to make our models more robust, we perform data augmentation operations on the input faces. 
In particular, we randomly apply downscaling, horizontal flipping, random brightness contrast, hue saturation, noise addition and finally JPEG compression.
Specifically, we resort to Albumentation \cite{albumentations} as our data-augmentation library, while we use Pytorch~\cite{pytorch} as Deep Learning framework. We train the models using Adam \cite{Kingma2014h} optimizer with hyperparameters equal to $\beta_1 = 0.9, \ \beta_2 = 0.999, \ \epsilon  = 10^{-8}$, and initial learning rate equal to $10^{-5}$. 

Independently from the used training strategy, given the size of the datasets, we never train our networks for a complete epoch. Specifically:
\begin{itemize}
    \item for the \textit{end-to-end training}, we either train for a maximum of $20\textrm{k}$ iterations, indicating as iteration the processing of a batch of $32$ faces ($16$ real, $16$ fake) taken randomly and evenly across all the videos of the train split, or until reaching a plateau on the validation loss. Validation of the model in this context is performed every $500$ training iterations, on $6000$ samples taken again evenly and randomly across all videos of the validation set. The initial learning rate is reduced of a $0.1$ factor if the validation loss does not decrease after $10$ validation routines ($5000$ training iterations), and the training is stopped when we reach a minimum learning rate of $1 \times 10^{-10}$;
    \item for the \textit{siamese training}, the feature extractor is trained using the same number of iterations, validation routine and learning rate scheduling of the end-to-end training. The main difference lies in the different loss function used (as explained in Section~\ref{sec:method}), and in the composition of the batch, which in this case is made by $12$ triplets of samples (6 real-fake-fake, 6 fake-fake-real) selected across all videos of the set considered. Regarding the parameter $\mu$ in \eqref{tripletloss}, we set it to $1$ after some preliminary experiments. The fine-tuning of the classification layer is then executed in a successive step following the end-to-end training paradigm with the hyperparameters specified above.
\end{itemize}
We finally run our experiments on a machine equipped with an Intel Xeon E5-2687W-v4 and a NVIDIA Titan V.
The code to replicate our tests is freely available at
\url{https://github.com/polimi-ispl/icpr2020dfdc}.

%% file: results.tex
\section{Results}\label{sec:results}
In this section we collect all the results obtained during our experimental campaign.

\begin{figure*}[t]
    \centering
    \includegraphics[width=\linewidth]{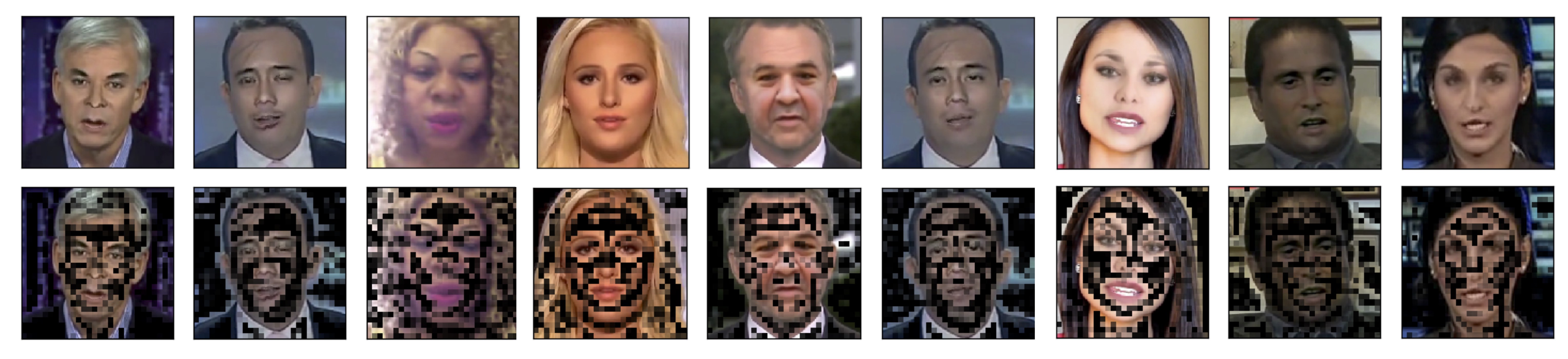}
    \caption{Effect of the attention on faces under analysis. Given some faces to analyze (top row), the attention network tends to select regions like eyes, mouth and nose (bottom row). Faces have been extracted from FF++ dataset.}
    \label{fig:faces_attention}
\end{figure*}

\subsection{EfficientNetB4Att explainability}

In order to show the effectiveness of the attention mechanism in extracting the most informative content of faces, we evaluate the attention map computed on a few faces of \gls{ff++}. 
Referring to Fig.~\ref{fig:efficientnetB4att}, we select the output of the Sigmoid layer in the attention block, which is a 2D map with size $28 \times 28$. Then, we up-scale it to the input face size ($224 \times 224$), and superimpose this to the input face. 
Results are reported in Fig. \ref{fig:faces_attention}. 
It is worth noting that this simple attention mechanism enables to highlight the most detailed portion of faces, e.g., eyes, mouth, nose and ears. 
On the contrary, flat regions (where gradients are small) are not informative for the network.
As a matter of fact, it has been shown several times that artifacts of deepfake generation methods are mostly localized around facial features \cite{verdoliva2020media}. For instance, roughly modeled eyes and teeth, showing excessively white regions, are still the main trademarks of these methods.


\subsection{Siamese features}

\begin{figure}[t]
    \centering
    \includegraphics[width=\columnwidth]{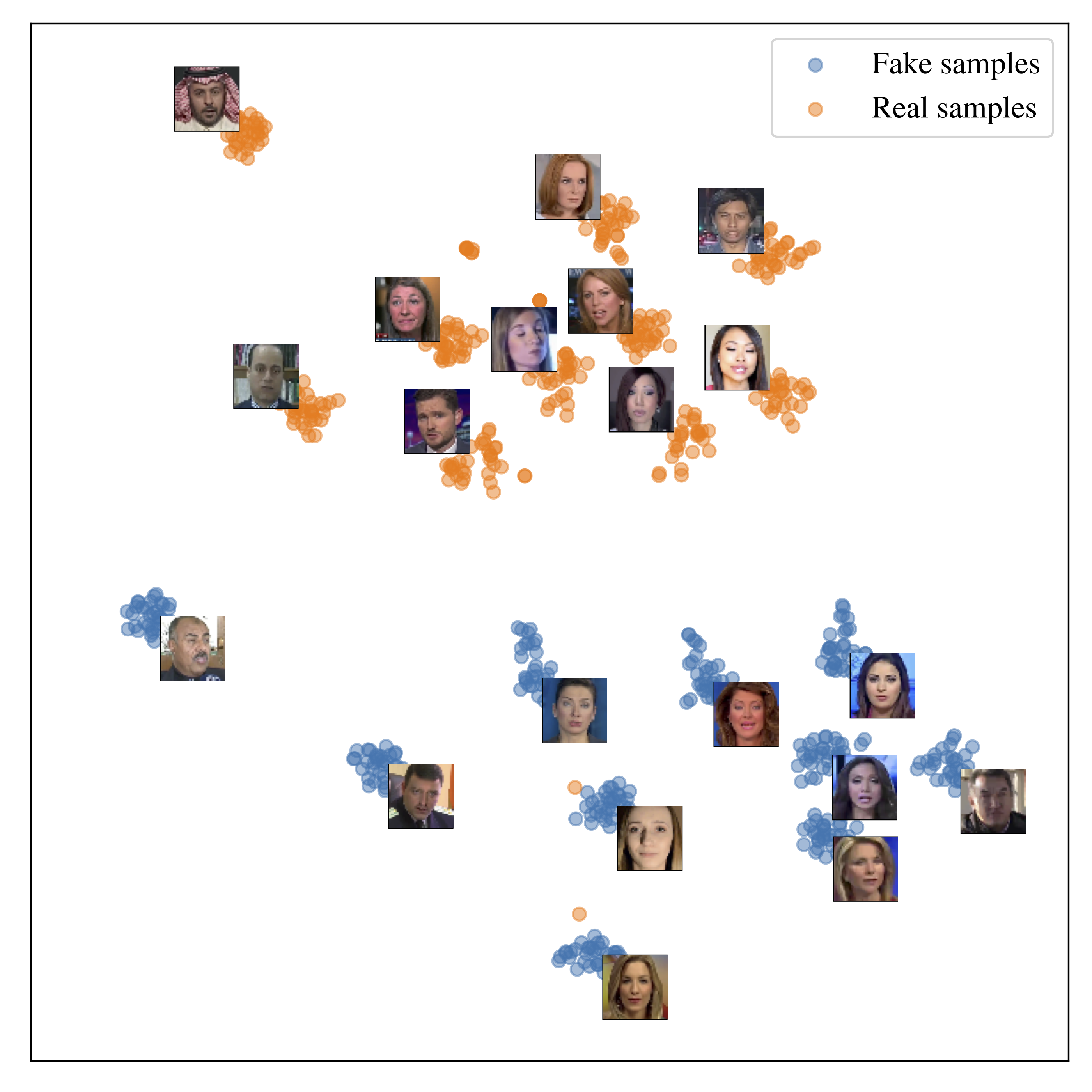}
    \caption{t-SNE visualization of features obtained by EfficientNetB4Att with siamese training. Faces have been extracted from FF++ dataset.}
    \label{fig:tsne}
\end{figure}

In order to understand whether the features produced by the encoding of the network when trained in siamese fashion are discriminatory for the task, we computed a projection over a reduced space using the well known algorithm t-SNE~\cite{vanDerMaaten2008tsne}.
In Fig.~\ref{fig:tsne} we show the projection obtained by means of EfficientNetB4Att starting from $20$ \gls{ff++} videos. 
We can clearly see how frames of the same videos clusters into small sub-regions. More importantly, all the real samples cluster into the top region of the chart, whereas the fake samples are in the bottom region. Frames of the same videos clusters into smaller sub-regions. This justifies the choice to adopt this particular training paradigm in addition to the classical end-to-end approach.

\subsection{Architecture independence}
As we want to understand whether the different networks can be used in an ensemble, we explore whether the scores extracted by each model are independent to some extent.

\begin{figure*}[t]
    \centering
    \subfloat[\gls{ff++}]{\includegraphics[width=\columnwidth]{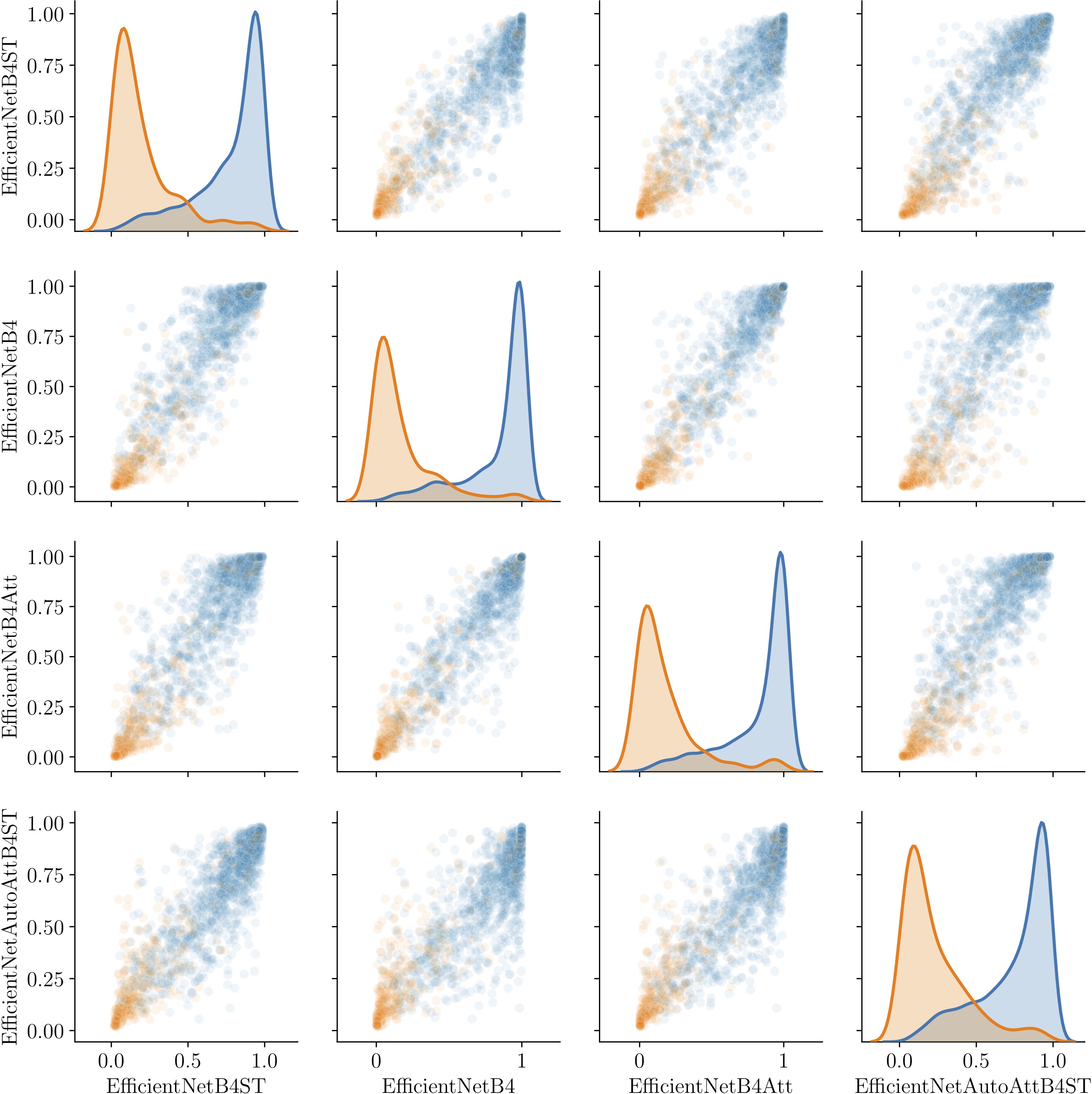}} \quad
    \subfloat[\gls{dfdc}]{\includegraphics[width=\columnwidth]{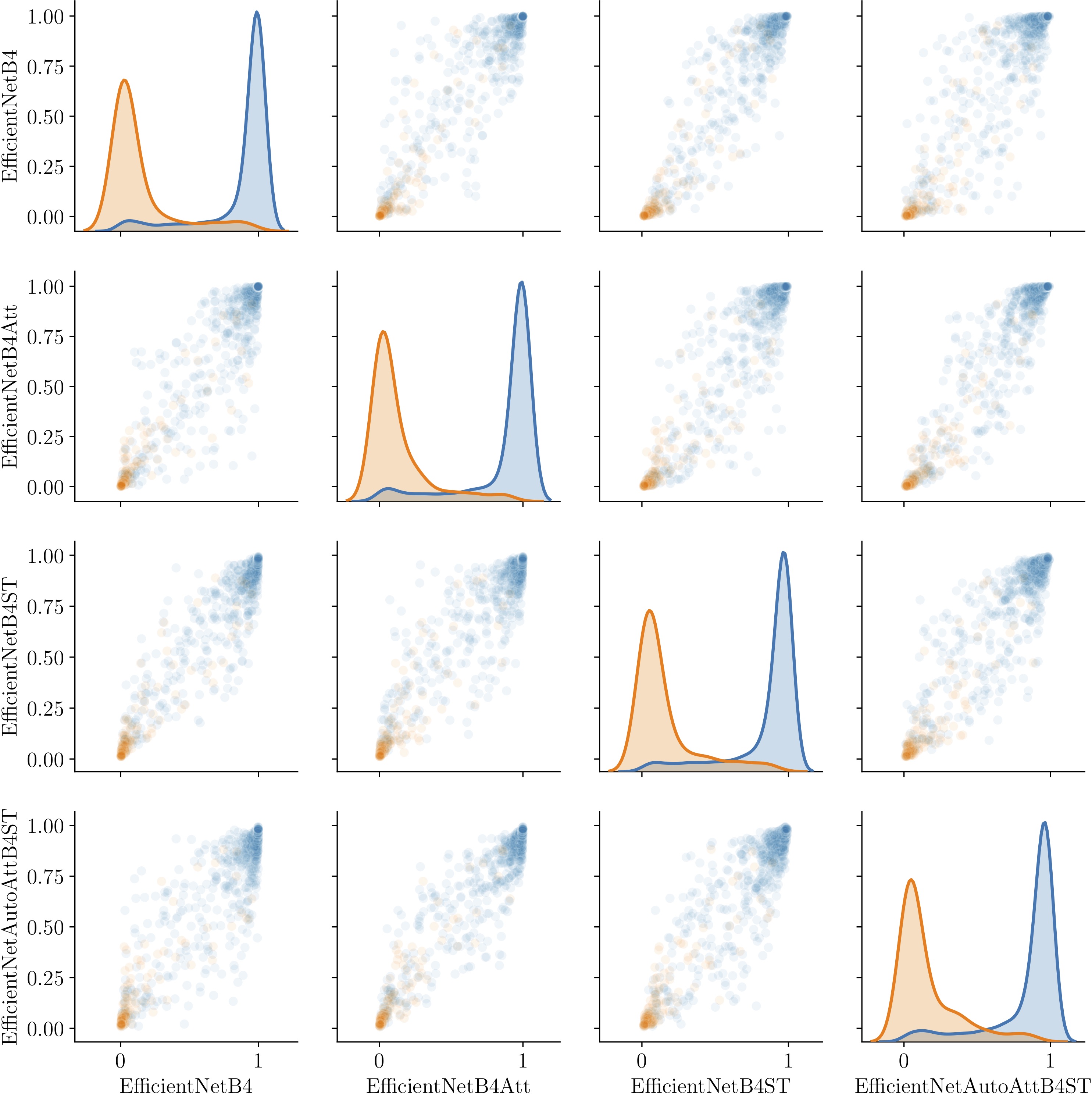}}
    \caption{Pair-plot showing the score distribution for real (orange \textcolor{mat_orange}{$\bullet$}) and fake (blue \textcolor{mat_blue}{$\bullet$}) samples for each pair of networks on \gls{ff++} (a) and \gls{dfdc} (b) datasets.}
    \label{fig:pairplot}
\end{figure*}

In Fig.~\ref{fig:pairplot}, all plots outside of the main diagonal show that different networks provide slightly different scores for each frame.
Indeed, the point clouds do not perfectly align on a shape that can be easily described by a simple relation.
This motivates us in using the different trained models in an ensemble way.
If all networks were perfectly correlated, this would not be reasonable.

\subsection{Face manipulation detection capability}
\begin{table}[t]
\centering
\caption{\gls{auc} and LogLoss obtained with different network combinations over all the datasets. Top-3 results per column in bold, baseline in italics.}
\label{tab:auc_loss}
\resizebox{\columnwidth}{!}{
\begin{tabular}{@{}ccccccccc@{}}
\toprule
\textbf{Xception} & \multicolumn{4}{c}{\textbf{EfficientNet}} & \multicolumn{2}{c}{\textbf{AUC}} & \multicolumn{2}{c}{\textbf{LogLoss}} \\
\textbf{Net} & \textbf{B4} & \textbf{B4ST} & \textbf{B4Att} & \textbf{B4AttST} & \textbf{\gls{ff++}} & \textbf{\gls{dfdc}}& \textbf{\gls{ff++}} & \textbf{\gls{dfdc}} \\ \midrule
\checkmark &  &  &  &  & \textit{0.9273} & \textit{0.8784} & \textit{0.3844} & \textit{0.4897}\\ \midrule
 & \checkmark &  &  &  & 0.9382 & 0.8766 & 0.3777 & 0.4819\\
 &  & \checkmark &  &  & 0.9337 & 0.8658 & 0.3439 & 0.5075\\
 &  &  & \checkmark &  & 0.9360 & 0.8642 & 0.3873 & 0.5133 \\
 &  &  &  & \checkmark & 0.9293 & 0.8360 & 0.3597 & 0.5507\\ \midrule
 & \checkmark & \checkmark &  &  & 0.9413 & \textbf{0.8800} & 0.3411 & 0.4687\\
 & \checkmark &  & \checkmark &  & 0.9428 & \textbf{0.8785} & 0.3566 & 0.4731\\
 & \checkmark &  &  & \checkmark & 0.9421 & 0.8729 & 0.3370 & 0.4739\\
 &  & \checkmark & \checkmark &  & 0.9423 & 0.8760 & 0.3371 & 0.4770\\
 &  & \checkmark &  & \checkmark & 0.9393 & 0.8642 & \textbf{0.3289} & 0.4977\\
 &  &  & \checkmark & \checkmark & 0.9390 & 0.8625 & 0.3515 & 0.4997\\ \midrule
 & \checkmark & \checkmark & \checkmark &  & \textbf{0.9441} & \textbf{0.8813} & 0.3371 & \textbf{0.4640}\\
 & \checkmark & \checkmark &  & \checkmark & 0.9432 & 0.8769 & \textbf{0.3269} & \textbf{0.4684}\\
 & \checkmark &  & \checkmark & \checkmark & \textbf{0.9433} & 0.8751 & 0.3399 & 0.4717\\
 &  & \checkmark & \checkmark & \checkmark & 0.9426 & 0.8719 & 0.3304 & 0.4800\\ \midrule
 & \checkmark & \checkmark & \checkmark & \checkmark & \textbf{0.9444} & 0.8782 & \textbf{0.3294} & \textbf{0.4658}\\ \bottomrule
\end{tabular}}
\end{table}

In this section, we report the average results achieved by the baseline network (i.e., XceptionNet) and the $4$ proposed models (i.e., EfficientNetB4, EfficientNetB4Att, EfficientNetB4ST and EfficientNetB4AttST). 
We also verify our guess behind the use of an ensemble, specifically combining two, three or even all the proposed models. 
In this case, the final score associated with a face is simply computed as the average between the scores returned by the single models.

In Table~\ref{tab:auc_loss} we report the \gls{auc} (computed binarizing the network output with different thresholds) and LogLoss obtained in our experiments.
Results are provided in a \textit{per-frame} fashion.

Analyzing these results, it is worth noting that the strategy of model ensembling generally awards in terms of performances. 
As somehow expected, best top-$3$ results are always reached by a combination of $2$ or more networks, meaning that network fusion helps both the accuracy of the deepfake detection (estimated by means of \gls{auc}) and the quality of the detection (estimated by means of LogLoss measure).
Indeed, on both datasets, LogLoss and AUC are always better than the baseline.

\subsection{Kaggle results}
In order to gain a deeper insight on the proposed solution performance, we also participated to the \gls{dfdc} challenge on Kaggle~\cite{dfdc2019} as \textit{ISPL} team.
The ultimate goal of the competition was to build a system able to tell whether a video is real or fake.
The \gls{dfdc} dataset used in this paper represents the training dataset released by the competition host, while the evaluation is performed over two different testing datasets: (i) the public test dataset; (ii) the private test dataset.
Participants were not aware of the composition of those datasets (e.g., the provenance of the sequences, the techniques used for generating fakes, etc.), apart from the number of videos in public test set, which is roughly $4000$.
The final solution proposed by our team was an ensemble of the $4$ proposed models, which led us to top $3\%$ on the leaderboard computed against the public test set. For the time being, the leaderboard computed over the private test set has not been disclosed yet.

%% file: conclusions.tex
\section{Conclusions}
\label{sec:conclusions}
Being able to detect whether a video contains manipulated content is nowadays of paramount importance, given the significant impact of videos in everyday life and in mass communications.
In this vein, we tackle the detection of facial manipulation in video sequences, targeting classical computer graphics as well as deep learning generated fake videos. 

The proposed method takes inspiration from the family of EfficientNet models and improves upon a recently proposed solution, investigating an ensemble of models trained using two main concepts: (i) an attention mechanism which generates a human comprehensible inference of the model, increasing the learning capability of the network at the same time; (ii) a triplet siamese training strategy which extracts deep features from data to achieve better classification performances.

Results evaluated over two publicly available datasets containing almost $120\,000$ videos reveals the proposed ensemble strategy as a valid solution for the goal of facial manipulation detection.

Future work will be devoted to the embedding of temporal information.
As a matter of fact, intelligent voting schemes when more frames are analyzed at once might lead to an increased accuracy.